\documentclass[10pt,twocolumn,letterpaper]{article}

\usepackage[pagenumbers]{cvpr} 

\usepackage{amsmath,amsfonts}
\usepackage{algorithmic}
\usepackage{algorithm}
\usepackage{array}
\usepackage{multirow}
\usepackage{amsmath,amsthm,amssymb}
\usepackage{color}

\newcommand{\beq}{\begin{equation}}
\newcommand{\eeq}{\end{equation}}
\newcommand{\beqs}{\begin{eqnarray}}
\newcommand{\eeqs}{\end{eqnarray}}
\newcommand{\barr}{\begin{array}}
	\newcommand{\earr}{\end{array}}
\newcommand{\bali}{\begin{aligned}}
	\newcommand{\eali}{\end{aligned}}

\newcommand{\Dc}[0]{\ensuremath{\mathcal{D}} }

\newcommand{\Lc}[0]{\ensuremath{\mathcal{L}} }

\newcommand{\Ebb}[0]{\ensuremath{\mathbb{E}} }

\newcommand{\Rbb}[0]{\ensuremath{\mathbb{R}} }

\newcommand{\st}[0]{\emph{s.t. }}

\newcommand{\Cmat}[0]{\ensuremath{{\bf C}} }

\newcommand{\tc}[0]{\ensuremath{\text{c}} }
\newcommand{\ts}[0]{\ensuremath{\text{s}} }

\newcommand{\sv}[0]{\ensuremath{\boldsymbol{s}} }

\newcommand{\uv}[0]{\ensuremath{\boldsymbol{u}} }

\newcommand{\wv}[0]{\ensuremath{\boldsymbol{w}} }
\newcommand{\xv}[0]{\ensuremath{\boldsymbol{x}} }

\newcommand{\zv}[0]{\ensuremath{\boldsymbol{z}} }

\newcommand{\thetav}[0]{\ensuremath{\boldsymbol{\theta}} }

\newcommand{\muv}[0]{\ensuremath{\boldsymbol{\mu}} }

\newcommand{\phiv}[0]{\ensuremath{\boldsymbol{\phi}} }

\newcommand{\KL}[0]{\ensuremath{\mathrm{KL}} }

\newcommand{\bb}[0]{\color{blue}}

\newcommand{\kk}[0]{\color{black}}

\usepackage{scalerel}
\newcommand\mathbox[1]{\mathord{\ThisStyle{%
			\fboxsep3\LMpt\relax\kern1\LMpt\fbox{$\SavedStyle#1$}\kern1\LMpt}}}

\definecolor{cvprblue}{rgb}{0.21,0.49,0.74}
\usepackage[pagebackref,breaklinks,colorlinks,allcolors=cvprblue]{hyperref}

\def\paperID{5922} 
\def\confName{CVPR}
\def\confYear{2026}

\title{Class-Conditional Distribution Balancing for \\ Group Robust Classification}

\author{Miaoyun Zhao\\
Dalian University of Technology\\
 Dalian China\\
{\tt\small  myzhao@dlut.edu.cn}
\and
Chenrong Li \\
Dalian University of Technology\\
Dalian China\\
{\tt\small 32409074@mail.dlut.edu.cn}
\and
Qiang Zhang\\
Dalian University of Technology\\
Dalian China\\
{\tt\small zhangq@dlut.edu.cn}
}

\begin{document}
\maketitle

\begin{abstract}
	Spurious correlations that lead models to correct predictions for the wrong reasons pose a critical challenge for robust real-world generalization.
	Existing research attributes this issue to group imbalance and addresses it by maximizing group-balanced or worst-group accuracy, which heavily relies on expensive bias annotations. 
	A compromise approach involves predicting bias information using extensively pretrained foundation models, which requires large-scale data and becomes impractical for resource-limited rare domains.
	To address these challenges, we offer a novel perspective by reframing the spurious correlations as imbalances or mismatches in class-conditional distributions, and propose a simple yet effective robust learning method that eliminates the need for both bias annotations and predictions.
	With the goal of maximizing the conditional entropy (uncertainty) of the label given spurious factors, our method leverages a sample reweighting strategy to achieve class-conditional distribution balancing, which automatically highlights minority groups and classes, effectively dismantling spurious correlations and producing a debiased data distribution for classification.
	Extensive experiments and analysis demonstrate that our approach consistently delivers state-of-the-art performance, rivaling methods that rely on bias supervision.
\end{abstract}

\section{Introduction}
\label{sec:intro}



Despite the unprecedented success of deep learning in recent decades, machine learning methods following the ERM paradigm \cite{vapnik1998statistical} often suffer from spurious correlations when training data is biased: some spurious factors (also termed bias) are predictive of the target class solely due to their high co-occurrence with the target label, despite being irrelevant to the true classification function \cite{labonte2023towards,geirhos2020shortcut}. 
It causes unpredictable performance losses under distribution shift, where spurious correlations in the test data differ from or conflict with those in the training data \cite{shah2020pitfalls,zhang2021deep,gulrajani2020search}.
For example, in the task of recognizing cows and camels \cite{tsirigotis2023group}, the training dataset is biased with camel images predominantly associating with desert backgrounds, leaving camels on grass as a minority group \cite{beery2018recognition}. 
Consequently, the classifier is likely to associate camels with deserts, resulting in a biased model that fails the underrepresented groups \cite{tatman2017gender}.
Moving beyond these simple instances, spurious correlations have surfaced as a major challenge in numerous high-stakes settings. This issue is especially evident in areas such as criminal justice systems \cite{wetherington2021spuriousness}, autonomous driving \cite{eykholt2018robust}, and medical diagnosis \cite{zech2018variable}, where ensuring fairness is of paramount importance \cite{caton2024fairness}. 

While progress has been made, existing research attributes spurious correlations to group imbalance.
They enhance robustness against spurious correlations by utilizing bias information to implement group balanced measures during training \cite{bahng2020learning,sagawadistributionally,kim2024discovering} or validation \cite{nam2020learning,liu2021just,liuavoiding,kirichenko2022last,zhou2022model}. 
These methods can be categorized into bias-supervised and pseudo-bias-supervised methods, depending on how bias information is obtained.
Bias-supervised methods \cite{bahng2020learning,sagawadistributionally,nam2020learning,liu2021just,liuavoiding,kirichenko2022last,zhou2022model} rely on ground-truth bias annotations provided by human annotators, which limits their applicability in practical scenarios where such annotations are too costly or even impossible to obtain. 
To address these limitations, pseudo-bias-supervised methods \cite{pezeshki2024discovering,asgari2022masktune,labonte2023towards,kim2024discovering,tsirigotis2023group} 
are developed to use predicted bias information instead of human annotations. By leveraging extensively pretrained base models \cite{he2020momentum} or large scale visual language models  \cite{mokady2021clipcap,li2022blip}, some methods achieve performance only slightly inferior to that of bias-supervised approaches. 
However, the pretraining process is both data- and resource-intensive, making it impractical for rare domains(\eg, hyperspectral images  \cite{scheibenreif2023masked}, medical images \cite{ma2024segment}, and point cloud data \cite{wu2024point}), where collecting large-scale paired vision-language datasets is impractical.

Here we consider a more realistic and challenging scenario where neither bias annotations nor a pretrained model for bias prediction are available.
With this goal, a line of research on stable learning has been proposed \cite{zhang2021deep,yu2023stable,kuang2020stable}. 
They make a structural assumption about data features, splitting them into core and spurious factors, and hypothesize that the statistical dependence between core and spurious factors is the primary cause of model failure under distribution shift. 
These methods propose a global sample reweighting strategy to remove spurious correlations by decorrelating all features, without requiring explicit bias information. However, with finite samples, achieving perfect independence is in theory challenging. Moreover, most of these methods are designed for linear frameworks with predefined features. When extended to deep models, however, their effectiveness diminishes due to the heavy entanglement of learned features, where mere decorrelation fails to ensure proper disentanglement \cite{yu2023stable}.

Motivated by these observations, 
we explore an orthogonal approach that maximizes the conditional entropy of the label given spurious factors through sample reweighting, enabling the model to make decisions free from spurious correlations.
We offer a novel perspective on the origins of spurious correlations: ($i$) distribution imbalance, referring to the class-conditional distribution mismatch, and ($ii$) class imbalance. 
Accordingly, we design a modified class-specific reweighting mechanism to address both types of imbalance. Building on this, we propose the class-conditional distribution balancing (CCDB) technique, which automatically reduces label predictability arising from spurious factors and thus yields a data distribution with spurious correlations removed. We further show theoretically that CCDB generalizes traditional class-balancing methods by incorporating conditional distribution matching, and that this matching is consistent with the ``covariate balance'' principle in causal inference \cite{neal2020introduction}. 
To further reinforce the mitigation of spurious correlations, we introduce a data-splitting strategy that deliberately exposes spurious factors more explicitly to the reweighting process.

The main contributions of this work are as follows:
	($i$) We design a class-specific sample reweighting mechanism and, based on it, a class-conditional distribution balancing technique that generalizes class balancing by adjusting the data from both class-conditional and class-marginal distributions.
	($ii$) By maximizing the entropy of the target label given spurious factors, we design a three-stage procedure to produce a reweighted data distribution that mitigates spurious correlations.
	($iii$) Our method saves the labor-intensive work of bias annotations or predictions. Extensive experiments on various synthetic and real-world datasets demonstrate its superiority over methods that rely on bias supervision.

\section{Related work}
\label{sec:relate}


\subsection{Bias-guided methods}

Addressing spurious correlations becomes straightforward with bias annotations. By combining both bias and label annotations, one can create a group-balanced training process (e.g., GroupDRO \cite{sagawadistributionally}) or model selection strategy (e.g., DFR \cite{kirichenko2022last}, MAPLE \cite{zhou2022model}) to enhance robustness against spurious correlations.
To reduce reliance on manual annotations, pseudo-bias-supervised methods propose using predicted bias information instead of explicit annotations, typically via auxiliary models, and focusing on data from the worst-performing groups (e.g., LFF \cite{nam2020learning}, LC \cite{liuavoiding}, DFA \cite{chu2021learning}, JTT \cite{liu2021just}). Additionally, recent research \cite{kim2024discovering, yu2024hallucidoctor} explores integrating large-scale vision-language models for more accurate bias prediction.
However, the performance of these methods depends on the prediction quality and still requires bias annotations during validation for optimal performance.

\subsection{Bias-agnostic methods}

To eliminate the dependence on bias annotations/predictions, researchers have developed bias-agnostic methods driven by different motivations.  

One line of work focuses on stable learning \cite{kuang2018stable},
which attributes model failure under distribution shifts to the statistical dependency between core and spurious features. They propose decorrelating all features to exclude spurious correlations \cite{kuang2020stable,zhang2021deep,yu2023stable}. 
However, most of these methods are developed within linear frameworks with predefined features, and their extension to deep models is less effective. This is because the learned features from deep models are often heavily entangled, making decorrelation insufficient to disentangle core and spurious factors \cite{yu2023stable}. 
Another line of research utilizes disagreements among diverse models to improve generalization under distribution shifts\cite{pagliardiniagree,lee2022diversify,labonte2023towards}, but these methods still require a small set of group annotations for model selection. 
Several other bias-agnostic methods have also demonstrated competitive performance
\cite{han2023general,tsirigotis2023group,pezeshki2024discovering,asgari2022masktune}.
However, most of these approaches may overestimate the level of bias, which necessitates the careful selection of hyperparameters to strike a better balance between majority and minority groups.

Our method also falls within the bias-agnostic framework. 
In contrast to the aforementioned methods, our approach ($i$) does not decorrelate all features, but instead takes a more straightforward approach by maximizing the conditional entropy of label information given the spurious factors;
($ii$) allows for easy control of sample weighting through distribution distance minimization.
Please refer to the appendix for a detailed discussion of related work.

\section{Our method}
\label{sec:our}

Given each observation consists of an input $\xv$ (\eg, an image $\xv\in\Rbb^{L\times H\times W}$) and a label $y\in\Rbb$, where $L$, $H$, and $W$ represent the channel, height and width of the image, respectively,
the spurious correlation problem is defined as follows.
Let $\zv\in\Rbb^{D}$ denote the latent representation of $\xv$ after feature extraction, which contains core and spurious factors, $\zv=[\zv_{\tc}, \zv_{\ts}]$. Here, $\zv_{\tc}\in\Rbb^{D_{\tc}}$ and $\zv_{\ts}\in\Rbb^{D_{\ts}}$ are implicitly entangled and technically challenging to decorrelate. $D$, $D_{\tc}$, and $D_{\ts}$ denote the dimensionalities of the corresponding latent variables.
Theoretically, $\zv_{\tc}$ represents the true core factors that define the class labels and is thus strongly correlated with $y$, whereas $\zv_{\ts}$ consists of spurious factors unrelated to class information.
However, under distribution shift, the presence of unknown confounders \cite{pearl2009causality} induces a strong spurious correlation between $\zv_{\ts}$ and $\zv_{\tc}$, and consequently between $\zv_{\ts}$ and $y$ in the training data \cite{christiansen2021causal}. 
This makes $\zv_{\ts}$ highly predictive of $y$ during training, but such a relationship does not hold at test time. Such unintended spurious correlations undermine model generalization. 
To address this issue, we break spurious correlations by reshaping the joint distribution $p(\zv_{\ts},y)$. Specifically, we make $\zv_{\ts}$ non-predictive of $y$ by maximizing the conditional uncertainty of $y$ given $\zv_{\ts}$, which corresponds to minimizing the negative conditional entropy as follows,
\begin{equation}\label{eq:mutual}
	\bali
	\Lc 
	&= - H(y|\zv_{\ts})
	= I(\zv_{\ts}, y) - H(y) \\
	&= \Ebb_{p(y)}\KL[p(\zv_{\ts}|y)\|p(\zv_{\ts})] + \Ebb_{p(y)}\log p(y)\\
	\eali
\end{equation} 
where $H(y|\zv_{\ts})$ is the conditional entropy of $y$ given $\zv_{\ts}$, $I(\zv_{\ts}, y)$ denotes mutual information between $\zv_{\ts}$ and $y$, $H(y)$ is the entropy of $y$, 
$\KL[p(\zv_{\ts}|y)\|p(\zv_{\ts})]$ is the Kullback–Leibler divergence \cite{kullback1951information}. 
Intuitively, \cref{eq:mutual} shows that spurious correlations arise from two primary factors: ($i$) the mismatch among the class-conditional distributions, termed distribution imbalance and ($ii$) the non-uniformity of the class distribution, termed class imbalance.
Motivated by these key insights, we propose the Class-conditional distribution balancing (CCDB) technique to break spurious correlations by minimizing \cref{eq:mutual}, which involves reducing the mutual information between $\zv_{\ts}$ and $y$ while maximizing the entropy of $y$.
Note that the first term in \cref{eq:mutual} drives the class-conditional distributions of $\zv_{\ts}$ toward a common marginal distribution. This can be interpreted as distribution-level balancing, which aligns with the notion of covariate balance in causal inference \cite{neal2020introduction,JMLR:v25:21-1028}.
Given the causal graph in \cref{fig:three_cases} (left), taking the binary classification as an example, treating $\zv_{\tc}$ as the treatment and $y$ as the outcome induces a causal path $\zv_{\tc}\rightarrow y$ as well as a backdoor path $\zv_{\tc} \leftarrow \uv \rightarrow \zv_{\ts} \rightarrow y$. 
The spurious correlation between $\zv_{\ts}$ and $y$ arises from the unknown confounder $\uv$. To safely estimate the causal effect of an image feature $\zv$ on the label $y$, one must remove the confounding bias induced by the differing distributions of $\zv_{\ts}$ between the treated and control image sets\footnote{Since $\uv$ is unknown, using $\zv_{\ts}$ provides an equivalent way to block the backdoor path.}. This requires enforcing $p(\zv_{\ts}|\zv_{\tc}) = p(\zv_{\ts})$, which is equivalent to our goal of achieving $p(\zv_{\ts}|y) = p(\zv_{\ts})$ in the KL term, given the one-to-one correspondence between $y$ and $\zv_{\tc}$.
The second term in \cref{eq:mutual} promotes uniformity across classes, ensuring class balancing.
Thus, \cref{eq:mutual} generalizes traditional class-balancing methods by addressing both distributional and class-level imbalances within a unified framework. 


Directly minimizing \cref{eq:mutual} is nontrivial because the disentangled $\zv_{\ts}$ and its exact distributions are not accessible. 
For practical implementation, we first propose learning a biased feature extractor in \cref{sec:pipline}, which captures features dominated by spurious correlations, \ie, $\zv \approx \zv_{\ts}$, and use these features to conduct spurious correlation mitigation. 
\textbf{Accordingly, to avoid cumbersome notation, we denote $\zv_{\ts}$ simply as $\zv$ in the following sections.}
Secondly, we introduce the following designations to handle each term in \cref{eq:mutual} separately.

\textbf{The KL term}. By approximating the data distribution with a Gaussian, we adopt the 2-Wasserstein distance \cite{mallasto2017learning,arjovsky2017wasserstein} as a substitute to measure the discrepancy between $p(\zv|y)$ and $p(\zv)$:
\begin{equation}\label{eq:wasser}
	W_2(\zv|_{y=k}, \zv) 
	= \|\muv_k-\overline{\muv}\|_2^2 + d(\Cmat_k,\overline{\Cmat}) \\
\end{equation}
where, $
d(\Cmat_k,\overline{\Cmat})=\text{trace}(\Cmat_k+\overline{\Cmat}-
2[\overline{\Cmat}^{\frac{1}{2}}\Cmat_k\overline{\Cmat}^{\frac{1}{2}}]^{\frac{1}{2}})$, $\{\muv_k, \Cmat_k\}$ and $\{\overline{\muv}, \overline{\Cmat}\}$ denote the mean and covariance matrix for $p(\zv|y=k)$ and $p(\zv)$, respectively.

\textbf{The entropy term}. This term can be easily maximized through data subsampling or reweighting, following traditional class-balancing techniques \cite{idrissi2022simple}. 
In fact, it is possible to handle these two terms together. 
In the following section, we propose a modified sample reweighting mechanism to achieve this goal.

\subsection{Intra-class sample reweighting for CCDB}

\begin{figure*}[t]
	\vspace{-0.5cm}
	\centering
	\includegraphics[width=1.99\columnwidth]{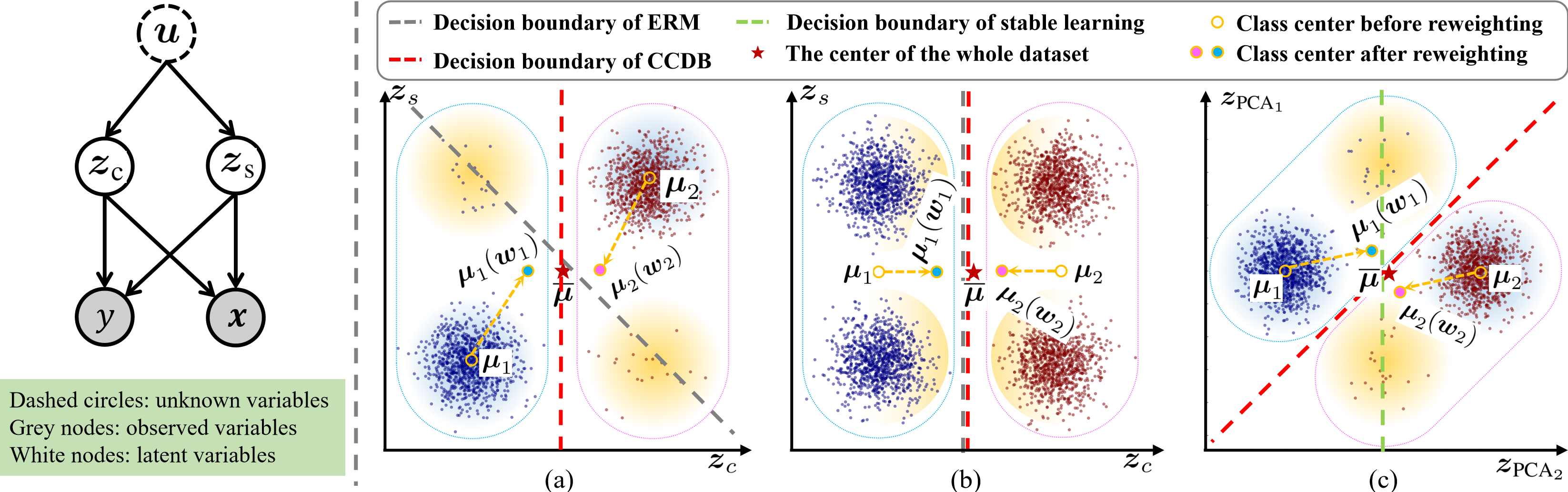}
	\vspace{-0.25cm}
	\caption{
		\textbf{Left:} Causal graph.
		\textbf{Right:} Analyzing the effectiveness of our method under three extreme configurations where $\zv_{\ts}$ and $y$ are: 
		(a) strongly correlated with well-disentangled predefined features;
		(b) completely independent; and
		(c) strongly correlated with predefined features that are independent but semantically entangled.
		Each case consists of two classes (marked in blue and red). 
		Data points in the yellow regions are assigned with high weights, while those in the blue regions are assigned with low weights.
	}\label{fig:three_cases}
	\vspace{-0.3cm}
\end{figure*}

Inspired by the success of sample reweighting in decorrelation \cite{zhang2021deep,yu2023stable}, we propose leveraging it to remove the spurious correlations between $\zv$ and $y$.
Unlike existing methods that learn a reweighting scheme over the entire training data, our approach applies sample reweighting independently within each class, which enables automatic optimization of both terms in \cref{eq:mutual}.

Suppose the observations are from $K$ classes, with each class represented as $\{\xv_i, y=k\}_{i=1}^{n_k}$, where $n_k$ denotes the sample count for class $k$.
We introduce class-specific sample weights to adjust the class-conditional distribution. Specifically, for class $k$, we define a weighting vector $\wv_k \in \Rbb^{n_k}$, satisfying $[\wv_k]_i\ge0$ for $i=1,\cdots,n_k$, and $\sum_{i=1}^{n_k}[\wv_k]_i=1$. 
After reweighting, the distribution distance between $p(\zv|y=k, \wv_k)$ and $p(\zv)$ following \cref{eq:wasser} is computed as:
\begin{equation}
	\bali
	W_2(\zv|\!_{\{y=k, \wv_{\!k}\!\}}\!,\zv) 
	\!=\! \|\muv_k(\wv_k)\!-\!\overline{\muv}\|_2^2
	\!+\! d(\Cmat_k(\wv_k),\overline{\Cmat})
	\eali
\end{equation}
where 
$\muv_k(\wv_k)$ and $\Cmat_k(\wv_k)$ denote the mean and covariance of class $k$ after reweighting, defined as, 
\vspace{-0.5cm}
\begin{equation}
	\bali
	\muv_k(\wv_k) &= 
	\sum_{i=1,y_i=k}^{n_k}[\wv_k]_i\cdot\zv_i\\
	\Cmat_k(\wv_k) &=\!\!\!\!\! \sum_{i=1,y_i=k}^{n_k}\!\!\!\![\wv_k]_i(\zv_i\!-\!\muv_k(\wv_k))^T(\zv_i\!-\!\muv_k(\wv_k))\\
	\eali
\end{equation}
\vspace{-0.1cm}

Note that we choose to set $\overline{\muv}$ and $\overline{\Cmat}$ as constants rather than functions of $\{\wv_1,\cdots,\wv_K\}$ for the following two reasons:
($i$) The marginal distribution $(\overline{\muv},\overline{\Cmat})$ serves as an anchor to stably guide the class-conditional distributions toward it. In the ideal case where all class-conditional distributions match the marginal distribution, the marginal distribution will remain unchanged.
($ii$) If, instead, $\overline{\muv}$ and $\overline{\Cmat}$ were functions of $\{\wv_1,\cdots,\wv_K\}$, the objective would become overly flexible and could lead to degenerate solutions—for example, allowing for many possible marginal distributions.

Accordingly, the objective in \cref{eq:mutual} is reformulated as follows,
\begin{equation}\label{eq:mu_loss}
	\Lc({\wv_1,\cdots,\wv_K}) = \frac{1}{K}\sum_{k=1}^{K}
	W_2(\zv|_{\{y=k, \wv_k\}}, \zv) 
\end{equation} 
where we omit the entropy term because $\wv_k$, being class-specific and summing to $1$, automatically assigns higher weights to classes with fewer samples, thus balancing the long-tailed class distribution.
The final objective for our CCDB is:
\begin{equation}\label{eq:final_loss}
	\bali
	\mathop{\min}_{\wv_1,\cdots,\wv_K}&\frac{1}{K}\sum_{k=1}^{K}
	W_2(\zv|_{\{y=k, \wv_k\}}, \zv) \\
	\st & \forall k\in\{1,\cdots,K\}, \sum_{i=1}^{n_k}[\wv_k]_i=1, \\
	& \forall i\in\{1,\cdots,n_k\}, [\wv_k]_i\ge0,
	\eali
\end{equation}

The optimal solution of \cref{eq:final_loss} reweights samples within each class by emphasizing those scarcely distributed in areas where the other classes are densely clustered, while understating samples in the opposite scenario.
This adjustment aligns the class-conditional distribution to the marginal distribution as closely as possible. In the ideal case,  $p(\zv|y=k, \wv_k)=p(\zv)$, indicating independence between $\zv$ and $y$.
Training a classifier on this reweighted dataset effectively removes reliance on spurious factors.

A toy example is provided in \cref{fig:three_cases}(right) to demonstrate the effectiveness of our CCDB in addressing three extreme situations where the spurious factors and class labels are either
strongly dependent (\cref{fig:three_cases}(a)(c)) or completely independent (\cref{fig:three_cases}(b)).  
In \cref{fig:three_cases}(a) and (c), $90\%$ of the samples have $\zv_{\ts}$ correlated with class labels, while the remaining $10\%$ display conflicting correlations. For each configuration, we draw the following conclusions:
\textbf{(a)} Since the data distribution strongly correlates $\zv_{\ts}$ with $y$, ERM places the decision boundary near the minority groups, leading to poor generalization in that region.
In contrast, our CCDB (and stable learning) emphasizes the minority area, resulting in a new distribution with the dependency between $\zv_{\ts}$ and $y$ significantly reduced. Thus the decision boundary is determined solely by $\zv_{\tc}$.
\textbf{(b)} When $\zv_{\ts}$ is independent of $y$, both ERM and CCDB correctly placed the decision boundary.
Additionally, CCDB further emphasizes samples near the boundary and improves performance in ambiguous regions.
\textbf{(c)} When predefined features are independent but each is a significant entanglement of core and spurious features (\eg, coincidentally aligned with a Principal Component Analysis (PCA) axis, with each direction being a mixture of $\zv_{\tc}$ and $\zv_{\ts}$), independent-based stable learning methods fail and place the decision boundary near the minority regions. In contrast, CCDB can still identify and emphasize minority groups, properly correcting the decision boundary.


\begin{algorithm}[t]
	\caption{
		CCDB
		for group robust classification
	}\label{alg:ccdb}
	\begin{algorithmic}
		\STATE {\bfseries Input:} Network $f_{\thetav_2}\circ f_{\thetav_1}$ for biased feature extraction, $f_{\phiv}$ for classification, logit vectors $\{\sv_c\}_{k=1}^{K}$ for sample reweighting,
		training set $\Dc_{tr}$ and validation set $\Dc_v$.
		\STATE {\bfseries Output:} unbiased classifier $f_{\phiv}$:
		\STATE  $\#$\bb \textit{Stage1: learning biased feature extractor}\kk
		\STATE \textbf{1:} Randomly split a subset from $\Dc_{tr}$ with proportion $\gamma$. 
		\STATE \textbf{2:} Train $f_{\thetav_2}\circ f_{\thetav_1}$ on the split subset using constrained ERM (\cref{eq:ERM_dist}).
		\STATE  $\#$ \bb\textit{Stage2: Learning sample weights}\kk
		\STATE \textbf{3:} Extract features $\{\zv_i\}_{i=1}^n$ from $\Dc_{tr}$ using the biased feature extractor $f_{\thetav_1}$.
		\STATE \textbf{4:} Obtain the optimal sample weights $\wv^*=\{\wv_1^*,\cdots,\wv_K^*\}$ by solving \cref{eq:final_loss}. 
		\STATE  $\#$ \bb\textit{Stage 3: Learning unbiased classifier}\kk
		\STATE \textbf{5:} Train classifier $f_{\phiv}$ on the reweighted samples using standard ERM. \\
		\STATE \textbf{6:} Select the best-performing $f_{\phiv}$ using $\Dc_v$.
	\end{algorithmic}
\end{algorithm}

\subsection{Full pipeline of our method}
\label{sec:pipline}

With the proposed reweighting mechanism, the complete pipeline of our CCDB is summarized in \cref{alg:ccdb}, which consists of three stages detailed as follows.

\begin{figure*}[!t]
	\vspace{-0.0cm}
	\setlength{\abovecaptionskip}{2.0pt}
	\centering
	\includegraphics[width=1.999\columnwidth]{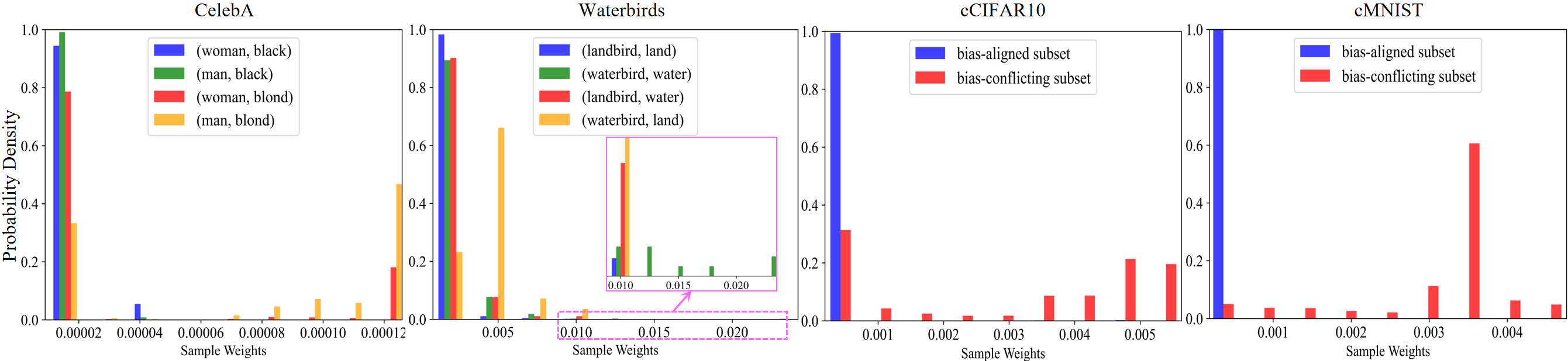}
	\vspace{-0.05cm}
	\caption{Distribution of the learned sample weights within each group. The distributions are normalized within each group. 
	}\label{fig:sample_weight}
	\vspace{-0.35cm}
\end{figure*}

\textbf{Stage 1: Learning biased feature extractor.} 
In this stage, we train a classifier $f_{\thetav_2}\circ f_{\thetav_1}$ on $\Dc_{tr}$, where $f_{\thetav_1}$ is the backbone and $f_{\thetav_2}$ is a fully connected classification head.
After training, $f_{\thetav_1}$ is used as a biased feature extractor in the next stage. 
Intuitively, we aim for the extracted features to be dominated by spurious factors as much as possible, so that their label predictability can be effectively reduced in the subsequent sample reweighting stage. 
To this end, we introduce the following two strategies:


\textit{Data splitting.} 
We experimentally find that the size of the training data significantly affects the dominance of spurious factors. 
As pointed out in \cite{pezeshki2024discovering}, when training on the entire training dataset, the model tends to overfit, causing the latent distribution to be less influenced by biased information. This, in turn, makes our method less effective.
\cref{fig:sample_weight} shows the latent space learned from different splits of the training data.
As the proportion of training data, $\gamma$, decreases, the extracted features become increasingly dominated by spurious factors. 
Based on this observation, we propose splitting the training data into a subset with proportion $\gamma$ (\eg, $\gamma=50\%$) and training $f_{\thetav_1}$ on it.
The resulting $f_{\thetav_1}$ is then applied to the entire training set for feature extraction and sample reweighting.
Our experiments show that data splitting is critical for fully exposing spurious factors, allowing them to be effectively balanced during the reweighting process.

\textit{Intra-class compactness.}
In addition to classification, we consider enhancing intra-class compactness in the latent space by minimizing the distance between samples belonging to the same class. 
The objective for feature extraction is thus reformed as follows:
\vspace{-5mm}
\begin{equation}\label{eq:ERM_dist}
	\bali
	&\Lc_{\thetav_1,\thetav_2} = \frac{1}{n}\sum_{i=1}^{n}l_\text{CE}(f_{\thetav_2}\circ f_{\thetav_1}(\xv_i), y_i) \\
	&+ \!
	\frac{\lambda}{K}\sum_{k=1}^{K}
	\frac{1}{n_k(n_k\!-\!1)}\!\!\sum_{i\neq j,y_i=y_j=k}
	\!\!\!\!\!\!\!\|f\!_{\thetav\!_1}(\xv_i)-f\!_{\thetav\!_1}(\xv_j)\|^2
	\eali
\end{equation}

where $l_{\text{CE}}$ denotes the cross entropy loss, $\|\cdot\|$ represents the Euclidean norm, $\lambda=0.5$ is used to control the strength of the constraint. 
In the presence of spurious correlations, certain spurious factors undoubtedly dominate within a specific class, acting as alternative features to $\zv_{\tc}$ for classification. Imposing intra-class compactness by bringing samples of the same class together serves two purposes (see appendix \cref{fig:gaussian}): ($i$) it clusters data with similar spurious factors, amplifying the bias attributes (samples with the same background tend to cluster together when intra-class compactness is applied); ($ii$) it smooths the feature space, making it resemble a Gaussian distribution, which is more favorable for the 2-Wasserstein distance we adopt.

\textbf{Stage 2: Learning sample weights.}
With the extracted biased features, we compute weights for each training sample by optimizing \cref{eq:final_loss}, which effectively reduces the dependence between spurious factors and class labels.
For practical implementation, we reparameterize the weight vector $\wv_k$ for class $k$ as the softmax of a logits vector $\sv_k\in \Rbb^{n_k}$, \ie, $\wv_k=\text{softmax}(\sv_k)$. This ensures that the weights are non-negative and sum to one, $\sum_{i=1}^{n_k}[\wv_k]_i=1$, with the elements of $\sv_k$ being real-valued. To stabilize optimization, we clip the elements of $\sv_k$ within the range $[-T, T]$, where the threshold $T=2$ typically works well.

\textbf{Stage 3: Learning unbiased classifier.}
Two approaches are available for incorporating the learned sample weights into standard ERM for unbiased learning.
($i$) Weighting samples during data sampling via \textit{``torch.utils.data.WeightedRandomSampler''}.
($ii$) incorporating the sample weights directly into the cross-entropy loss.
We adopt the first approach, as it empirically yields better performance.

\begin{table*}[h!]
	\vspace{-0.5cm}
	\centering
	\caption{
		Results on cMNIST and cCIFAR10 with various bias-conflicting ratios in the training set. The test accuracy(\%) is averaged over 5 random seeds.
		The best results are indicated in bold.
		\label{tab:CMNIST}
	}
	\vspace{-0.3cm}
	\renewcommand{\arraystretch}{1.0}
	\resizebox{0.98\hsize}{!}{
		\setlength{\tabcolsep}{4pt}
		\begin{tabular}{c|cc|cccc|cccc}
			\toprule
			\multirow{2}{*}{Methods}& \multicolumn{2}{c|}{Bias label} & \multicolumn{4}{c|}{cMNIST} & \multicolumn{4}{c}{cCIFAR10}\\
			\cline{2-11}
			&   Train & Val & $0.5\%$ & $1\%$ & $2\%$ & $5\%$ & $0.5\%$ & $1\%$ & $2\%$ & $5\%$\\
			\hline 
			GroupDRO & Yes & Yes & 63.12 & 68.78 & 76.30& 84.20 & 
			33.44 & 38.30 & 45.81 & 57.32  \\
			LFF & No & Yes & 52.50\scriptsize{$\pm$2.43} & 61.89\scriptsize{$\pm$4.97} & 71.03\scriptsize{$\pm$2.44} & 80.57\scriptsize{$\pm$3.84} & 
			28.57\scriptsize{$\pm$1.30} & 33.07\scriptsize{$\pm$0.77} & 39.91\scriptsize{$\pm$0.30} & 50.27\scriptsize{$\pm$1.56}  \\
			LC & No & Yes & 71.25\scriptsize{$\pm$3.17} & 82.25\scriptsize{$\pm$2.11} & 86.21\scriptsize{$\pm$1.02} & 91.16\scriptsize{$\pm$0.97} &
			34.56\scriptsize{$\pm$0.69} & 37.34\scriptsize{$\pm$0.69} & 47.81\scriptsize{$\pm$2.00} & 54.55\scriptsize{$\pm$1.26}  \\
			\hline 
			ERM & No & No & 35.19\scriptsize{$\pm$3.49} & 52.09\scriptsize{$\pm$2.88} & 65.86\scriptsize{$\pm$3.59} & 82.17\scriptsize{$\pm$0.74}& 23.08\scriptsize{$\pm$1.25} & 25.82\scriptsize{$\pm$0.33} & 30.06\scriptsize{$\pm$0.71} & 39.42\scriptsize{$\pm$0.64}  \\
			uLA & No & No & 75.13\scriptsize{$\pm$0.78} & 81.80\scriptsize{$\pm$1.41} & 84.79\scriptsize{$\pm$1.10} & 92.79\scriptsize{$\pm$0.85} & 
			34.39\scriptsize{$\pm$1.14} & 62.49\scriptsize{$\pm$0.74} & 63.88\scriptsize{$\pm$1.07} & 74.49\scriptsize{$\pm$0.58} \\
			\textbf{CCDB} & No & No & \textbf{83.20\scriptsize{$\pm$2.17}} & \textbf{87.95\scriptsize{$\pm$1.59}} & \textbf{91.02\scriptsize{$\pm$0.28}} & \textbf{96.37\scriptsize{$\pm$0.25}} & 
			\textbf{55.07\scriptsize{$\pm$0.85}} & \textbf{63.28\scriptsize{$\pm$0.46}} & \textbf{67.78\scriptsize{$\pm$0.78}} & \textbf{74.64\scriptsize{$\pm$0.34}} \\
			\bottomrule
		\end{tabular}
	}
	\vspace{-0.2cm}
\end{table*}

\begin{table*}[ht]
	\centering
	\caption{
		Results on Waterbirds, CelebA and CivilComments. We report average test accuracy(\%) and std.dev. over 5 random seeds.
		\label{tab:waterbirds}
	}
	\vspace{-0.3cm}
	\renewcommand{\arraystretch}{0.9}
	\resizebox{0.99\hsize}{!}{
		\setlength{\tabcolsep}{8pt}
		\begin{tabular}{c|cc|cc|cc|cc}
			\toprule
			\multirow{2}{*}{Methods}& \multicolumn{2}{c|}{Bias label} & \multicolumn{2}{c|}{Waterbirds} & \multicolumn{2}{c|}{CelebA}&  \multicolumn{2}{c}{CivilComments}\\
			\cline{2-9}
			&   Train & Val & i.i.d. & WorstGroup & i.i.d. & WorstGroup & i.i.d. & WorstGroup \\
			\hline 
			GroupDRO & Yes & Yes & 93.50 & 91.40 & 92.90 & 88.90 & 84.2 & 73.7 \\
			LFF & No & Yes & 97.50 & 75.20  & 86.00 & 77.20 & 68.2 & 50.3\\
			JTT & No & Yes & 93.60 & 86.00  & 88.00 & 81.10 & 83.3 & 64.3\\
			LC & No & Yes & - & 90.50\scriptsize{$\pm$1.1}  & - & 88.10\scriptsize{$\pm$0.8} & - & 70.30\scriptsize{$\pm$1.2}\\
			SELF & No & Yes & - & 93.00\scriptsize{$\pm$0.3}  & - & 83.90\scriptsize{$\pm$0.9} & - & 79.10\scriptsize{$\pm$2.1}\\
			\hline 
			ERM & No & No & \textbf{97.30} & 72.60 &\textbf{ 95.60} & 47.20 & 81.6 & 66.7\\ 
			MASKTUNE & No & No & 93.00\scriptsize{$\pm$0.7} & 86.40\scriptsize{$\pm$1.9}& 91.30\scriptsize{$\pm$0.1} &  78.00\scriptsize{$\pm$1.2} & - & -\\
			uLA & No & No & 91.50\scriptsize{$\pm$0.7} & 86.10\scriptsize{$\pm$1.5}& 93.90\scriptsize{$\pm$0.2} &  86.50\scriptsize{$\pm$3.7} & - & -\\ 
			XRM & No & No & 90.60 & 86.10 & 91.0 & \textbf{88.5} & 83.5 & 70.1\\ 
			\textbf{CCDB} & No & No & 92.59\scriptsize{$\pm$0.10}  & \textbf{90.48}\scriptsize{$\pm$0.28} & {90.08\scriptsize{$\pm$0.19}} & 85.27\scriptsize{$\pm$0.28} & \textbf{83.60}\scriptsize{$\pm$0.21} & \textbf{75.00}\scriptsize{$\pm$0.26} \\
			\bottomrule
		\end{tabular}
	}
	\vspace{-0.2cm}
\end{table*}

\section{Experimental results}
\label{sec:experiments}

\textbf{Dataset.}  
We conduct experiments on five widely used benchmark datasets: 
cMNIST \cite{li2022invariant}, 
cCIFAR10 \cite{hendrycks2018benchmarking},
Waterbirds \cite{zhou2022model}, 
CelebA \cite{sagawadistributionally}, 
and CivilComments \cite{koh2021wilds} 
to demonstrate the effectiveness of our method.
Among them, cMNIST and cCIFAR10 are ten-way classification tasks, where each class is dominated by a specific color or noise pattern, making them ideal for demonstrating the superior performance of CCDB in challenging multi-class settings with extremely biased training data.
Waterbirds, CelebA, and CivilComments are real-world datasets: Waterbirds and CelebA are image classification datasets where each class is strongly correlated with a specific background or gender attribute; CivilComments is a text classification dataset with the class information spuriously
correlated with mentions of certain demographic identities. 
These datasets are used to showcase the generalizability of CCDB to real-world scenarios. 
Detailed data information is provided in the supplementary material.
For all datasets, we use the standard train-validation-test splits as defined in their respective literature and report both group-balanced test accuracy and the worst-group test accuracy.

\textbf{Compared baselines.}
To demonstrate the superiority of our CCDB in addressing distribution shift in the presence of spurious correlations, we compare it with
a diverse set of group robustness techniques. GroupDRO \cite{sagawadistributionally} is bias-supervised during both training and validation, serving as a challenging baseline. 
LFF \cite{nam2020learning}, JTT \cite{liu2021just}, LC \cite{liuavoiding},
and SELF \cite{labonte2023towards} are pseudo-bias-supervised during training but require bias annotations during validation to achieve optimal performance.
Similar to our method, standard empirical risk minimization(ERM), uLA \cite{tsirigotis2023group}, MASKTUNE \cite{asgari2022masktune}, and XRM \cite{pezeshki2024discovering} are bias-agnostic during both training and validation.

\textbf{Training setup.}
To make a fair comparison, we apply a 3-hidden layer MLP for cMNIST, a ResNet18 for cCIFAR10, a ResNet50 for Waterbirds and CelebA, and a BERT for CivilComments following existing methods \cite{tsirigotis2023group}. 
Both ResNet18 and ResNet 50 are pretrained on ImageNet-1K, BERT is pretrained on Book Corpus and English Wikipedia.
The proposed CCDB method consists of three stages, each optimized independently. The first two training stages span 20 epochs and 1000 iterations across all datasets (except CivilComments with 10 epochs); the third stage involves 2000 iterations for Waterbirds and CelebA, 5000 iterations for cMNIST and cCIFAR10, and 10 epochs for CivilComments. 


On data preprocessing, we apply simple augmentations to each dataset following \cite{ahuja2021invariance}, which are designed not to induce invariance to variations in the bias attribute. Consequently, the improvements reported in the main paper can be attributed to the proposed CCDB methodology, rather than to the use of bias-nullifying data augmentations.
All the results are averaged over 5 random seeds.
See supplementary material for the detailed experimental setup.

\begin{figure*}[!t]
	\vspace{-0.5cm}
	\setlength{\abovecaptionskip}{2.0pt}
	\centering
	\includegraphics[width=1.999\columnwidth]{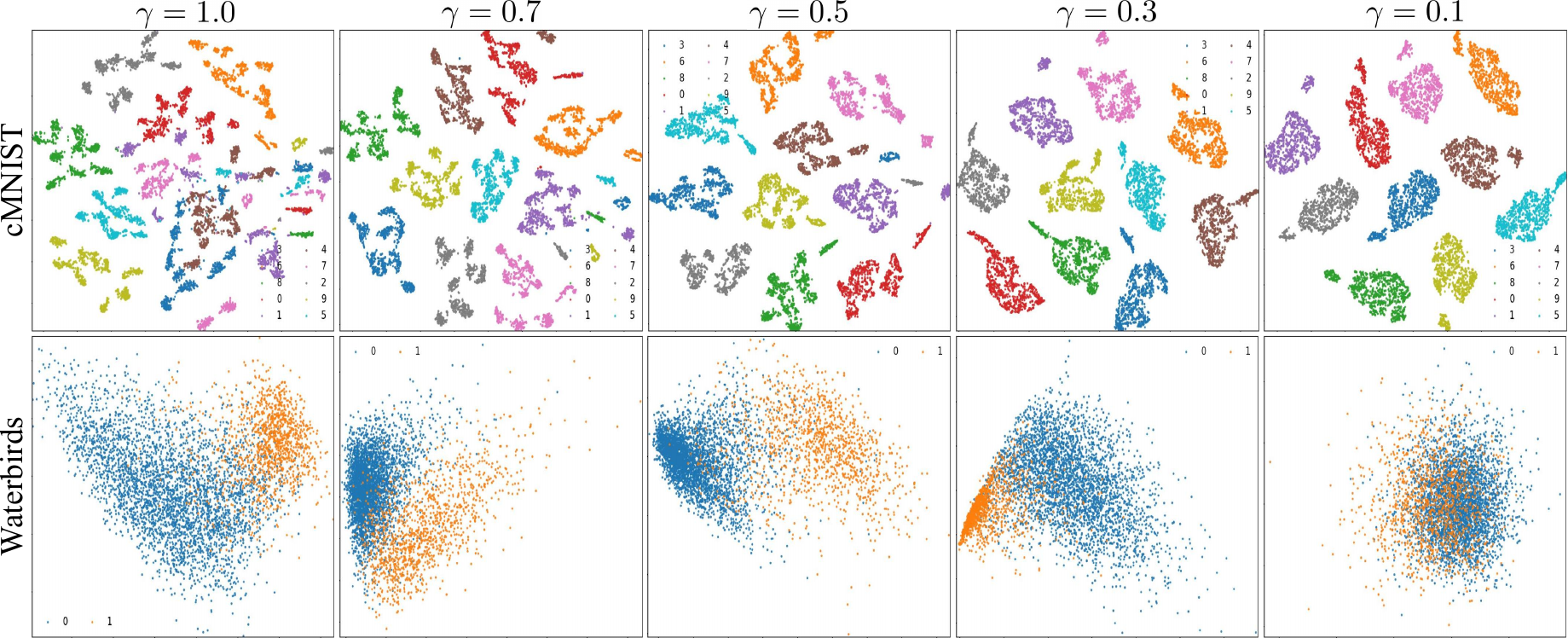}
	\caption{t-SNE visualization of the latent space learned in Stage1. Colored by bias annotations. 
		As $\gamma$ decreases, the feature space exhibits an increasingly pronounced clustering effect aligned with bias annotations. 
	}\label{fig:tsne}
	\vspace{-0.4cm}
\end{figure*}

\subsection{Experiments on multi-class data}

In this section, we use multi-class datasets (i.e., cMNIST and cCIFAR10) with controllable synthetic bias to evaluate the effectiveness of our method in mitigating varying levels of bias under challenging multi-class conditions.
For each dataset, we control the degree of spurious correlation by varying the proportion of bias-conflicting samples in the training data, and evaluate performance on completely unbiased test set.
Following \cite{tsirigotis2023group}, the bias-conflicting ratios are set to $\{0.5\%,1\%,2\%,5\%\}$ for both datasets, with $0.5\%$ representing an extremely biased scenario.
The results are reported in \cref{tab:CMNIST}. From which, we have the following observations: 
($i$) On cMNIST, our method consistently achieves the best performance, surpassing the second best with a large margin, even though some competing methods utilize bias annotations during model training or selection; 
($ii$) On cCIFAR10, our method performs the best on average and significantly outperforms the others when the bias-conflicting ratio is as low as $0.5\%$. Considering that uLA employs resource-consuming MOCOV2+ \cite{he2020momentum,chen2020improved} pretraining and broad hyperparameter searching, our method provides a favorable trade-off between performance and computational complexity.

\subsection{Experiments on real-world data}

Experiments on real-world datasets, Waterbirds, CelebA, and CivilComments are conducted to
test the effectiveness of our method in dealing with spurious correlations in practical scenarios. 
The results are reported in \cref{tab:waterbirds}. 
With the supervision of bias annotations, GroupDRO demonstrates strong generalization performance on the worst group.
On all three datasets, our method achieves competitive or superior performance compared to those utilizing bias information during validation.
Despite employing a simpler learning process, our method performs comparably to uLA on CelebA, and significantly surpasses uLA and XRM on Waterbirds and CivilComments.
By referring to \cref{tab:CMNIST}, we observe that bias supervision is more effective for simple binary classification tasks, \eg, GroupDRO achieves SOTA performance on Waterbirds and CelebA. However, for more complex multi-class classification tasks, bias-supervised methods fall significantly behind bias-free approaches(\eg, uLA and our CCDB).

\begin{figure*}[t]
	\vspace{-0.5cm}
	\setlength{\abovecaptionskip}{2.0pt}
	\centering
	\includegraphics[width=1.9\columnwidth]{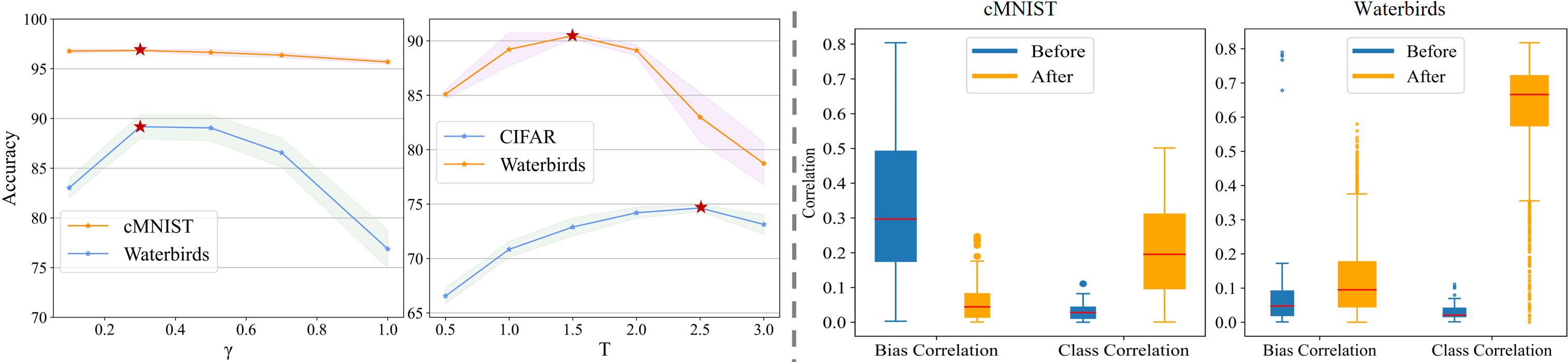}
	\vspace{-0.05cm}
	\caption{\textbf{Left:} Ablation study on the split ratio $\gamma$ and threshold $T$.
	\textbf{Right:} Boxplot of the feature correlations with bias and class information, before and after CCDB sample reweighting. Refer to appendix for more results.
	}\label{fig:split_clip}
	\vspace{-0.2cm}
\end{figure*}

\begin{figure*}[t]
	\vspace{-0.2cm}
	\setlength{\abovecaptionskip}{2.0pt}
	\centering
	\includegraphics[width=1.999\columnwidth]{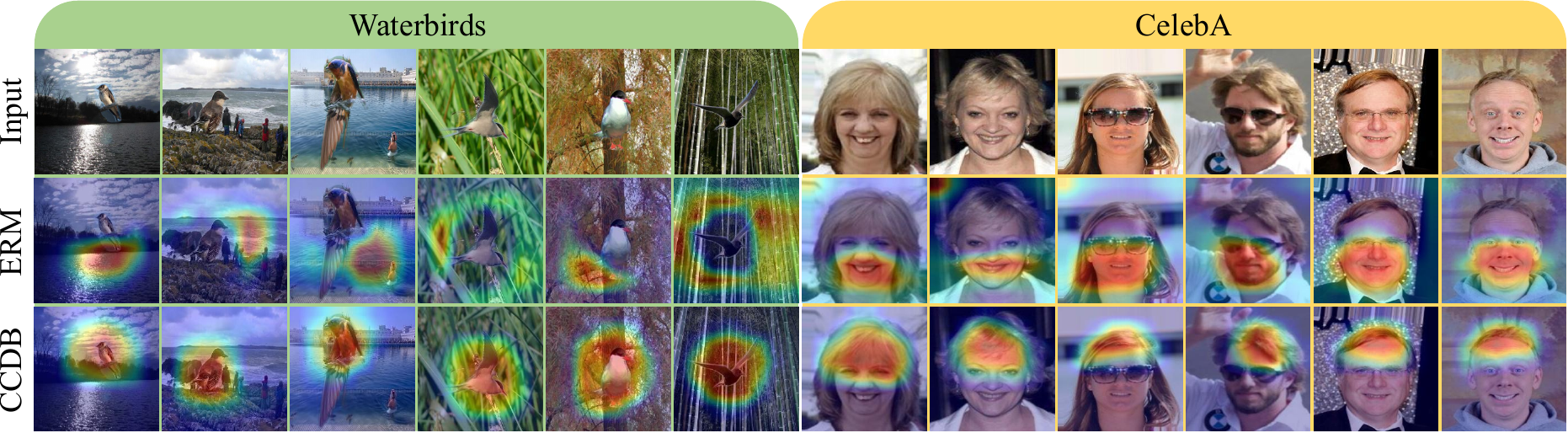}
	\vspace{-0.05cm}
	\caption{Visualization of ERM and CCDB attention on Waterbirds and CelebA samples. 
	}\label{fig:attention}
	\vspace{-0.5cm}
\end{figure*}

\subsection{Ablation study}

To guide the configuration of our method, we conduct an ablation study to investigate the impact of the following two factors on the final performance.

\textbf{The proportion $\gamma$ split from the training data for biased feature extractor learning.}
By varying $\gamma$ in $\{10\%,30\%,50\%,70\%,100\%\}$, we visualize the t-SNE of the learned feature space during stage1 in \cref{fig:tsne} and report the worst group accuracy of stage 3 in \cref{fig:split_clip} (left). 
In \cref{fig:tsne}, for cMNIST, as $\gamma$ decreases, the feature space exhibits an increasingly pronounced clustering effect aligned with bias annotations, indicating a feature space dominated by spurious factors.
For Waterbirds, the dominance of spurious factors initially increases but then diminishes when $\gamma=0.1$, likely due to the smaller scale of the training data.
Accordingly, with an appropriately small $\gamma\in[0.3,0.5]$, spurious factors are more effectively identified and balanced, leading to a classifier that relies more on core factors and exhibiting improved generalization capabilities, see the optimal test accuracy (marked with a red star) in \cref{fig:split_clip} (left) for verification.



\textbf{The threshold $T$ for sample weight clipping.}
The threshold $T$ controls the extent to which minority groups are emphasized, with larger values assigning greater weight to samples from minority groups. 
We evaluate the impact of $T$ by varying it from $0.5$ to $3.0$ and report the classification results in \cref{fig:split_clip}(left).
For datasets that are relatively less biased, \eg, Waterbirds, a threshold around $1.5$ yields the best performance. 
In contrast, for highly biased datasets, \eg, cCIFAR10 with a bias-conflicting ratio below $5\%$, a higher threshold around $2.5$ proves more effective.

\subsection{Analysis on sample weights learned by CCDB}

To show that the proposed sample reweighting technique effectively reduces the correlation between spurious factors $\zv_{\ts}$ and class labels $y$, we analyze the distribution of the learned sample weights and their impact on the spuriousness of the features extracted for classification.
\cref{fig:sample_weight} shows the histograms of sample weights obtained by our CCDB. 
For CelebA and Waterbirds, the minority groups (man-blond, waterbird-land, landbird-water) receive weights roughly ten times higher than the majority groups, reflecting our primal goal of up-weighting underrepresented samples. 
For cCIFAR10 and cMNIST, we divide the data into a bias-aligned subset, where $\zv_{\ts}$ aligns with $y$, and a bias-conflicting subset, where $\zv_{\ts}$ conflicts with $y$.
Obviously, the weights for the minority conflicting subset are concentrated around relatively high values, while the aligned subset has weights close to zero.

\cref{fig:split_clip}(right) shows the effect of sample reweighting on the spuriousness of the learned features. 
We measure spuriousness by calculating the Pearson correlation coefficient \cite{Benesty2009} between each feature dimension and class/bias information, and present a boxplot of these coefficients. 
Before reweighting, most extracted feature dimensions are dominated by spurious factors, showing high correlation with bias and low correlation with labels (blue boxes).
After CCDB reweighting, the features become predominantly core, strongly correlated with class labels and weakly correlated with bias (yellow box).
These results highlight the effectiveness of CCDB in mitigating spurious correlations, yielding a de-biased data distribution for downstream tasks.

To further demonstrate the effectiveness of the proposed CCDB in getting rid of spurious correlations, we compare the classifier's attention before and after sample reweighting in \cref{fig:attention} (visualized via xGradCAM \cite{selvaraju2017grad}). 
Without sample reweighting, the model obtained from ERM primarily focuses on spurious factors, \eg, background or facial details. In contrast, after applying CCDB reweighting, the model is enforced to explore core factors, \eg, the foreground birds and hair color, leading to more robust predictions.

\section{Conclusion}

In this work, we considered the challenge problem of preventing models from learning spurious correlations without relying on bias annotations. 
We provide a new perspective on the cause of spurious correlations: the imbalance in both class-conditional and class-marginal distributions. Accordingly, we propose a novel class-conditional distribution balancing (CCDB) technique that
effectively removes spurious correlations from the training data via a modified class-specific sample reweighting mechanism. 
The proposed CCDB performs exceptionally well in both basic binary and challenging multi-class scenarios without relying on bias/group predictions or expensive pretrained foundation models, offering a novel and promising direction for mitigating spurious correlations. 


{
    \small
    \bibliographystyle{unsrt}
    \bibliography{ReferencesCong}
}


\clearpage
\setcounter{page}{1}
\maketitlesupplementary

\section{Related works}

We categorize prior literature according to whether bias information is used to mitigate spurious correlations, as detailed below.

\subsection{Bias-guided methods}

Addressing spurious correlations becomes straightforward when bias annotations are available. 
By combining both bias and label annotations, one can create a group balanced training process or model selection strategy, to enhance robustness against spurious correlations.
For example, Group distributionally robust optimization (GroupDRO) \cite{sagawadistributionally,duchi2021statistics,oren2019distributionally,hu2018does} proposes to minimize the worst-group loss based on predefined group assignments.
To alleviate the reliance on bias annotations,
deep feature reweighting (DFR) \cite{kirichenko2022last,izmailov2022feature} proposes retraining only the last layer of the model using a small, group-balanced held-out dataset.
MAPLE \cite{zhou2022model} utilizes a validation set with bias annotations to reweight the training samples, ensuring that the training process focuses on the worst group. 
Pseudo-bias-supervised methods propose to use the predicted bias information instead of explicit annotations, typically through auxiliary models. Learning from Failure (LFF) \cite{nam2020learning}, logit adjustment technique (LC) \cite{liuavoiding} and Disentangled Feature Augmentation (DFA) \cite{chu2021learning} reweight the loss of the unbiased model using a co-trained biased model. Meanwhile, Just-train-twice (JTT) \cite{liu2021just} reweights misclassified samples from an initial biased training phase, emphasizing data from the worst-performing groups. Additionally, research in \cite{kim2024discovering,yu2024hallucidoctor,clark2019don} integrates large-scale vision-language models for more precise bias prediction.
However, the performance of these methods depends on the prediction quality and still requires bias annotations during validation for optimal performance.

\begin{table*}[!t]
	\centering
	\caption{
		An illustration of the datasets used in this paper. In the training data, $\zv_{\text{s}}$ is highly correlated with the class labels. However, this correlation is either broken or even reversed in the test set.
	}
	\label{tab:data_setting}
	\renewcommand{\arraystretch}{1.4}
	\resizebox{1.0\hsize}{!}{
		\setlength{\tabcolsep}{2pt}
		\begin{tabular}{c|cc|cccc}
			\toprule
			Dataset & $\zv_{\tc}$ & $\zv_{\ts}$ & classes & Train & Val & Test \\
			\midrule
			cMNIST & digit & digit color & 10 & ${95\%\sim99.5\%}$ biased & ${95\%\sim99.5\%}$ biased & unbiased \\
			cCIFAR10 & object & noise & 10 & ${95\%\sim99.5\%}$ biased & ${95\%\sim99.5\%}$ biased & unbiased \\
			Waterbirds & bird & background & 2 & $\{3498, 184, 56, 1057\}$ & \{467,467,133,133\} & \{2225,2225,642,642\} \\
			CelebA & hair color & gender & 2 & $\{71629,66874,22880,1387\}$ & \{8535,8276,2874,182\} & \{9767,7535,2480,180\}  \\
			CivilComments & toxic & identity & 2 & $\{148186,90337,12731,17784\}$ & \{25159,14966,2111,2944\} & \{74780,43778,6455,8769\}  \\
			\bottomrule
		\end{tabular}
	}
\end{table*}

\begin{table*}[!t]
	\centering
	\caption{
		The hyperparameters and augmentations used during the training of our CCDB. 
	}
	\label{tab:augmentation}
	\renewcommand{\arraystretch}{1.4}
	\resizebox{1.0\hsize}{!}{
		\setlength{\tabcolsep}{3pt}
		\begin{tabular}{c|cccccccc}
			\toprule
			Dataset & Learning rate & Batch size & Weight decay & $\{\text{Epoch,Iter,Iter}\}$ & $\lambda$& Split$\gamma$ & Clip$T$ & Augmentation  \\
			\midrule
			cMNIST &1e-2&256&1e-4&$\{5,1000,5000\}$&0.5& 0.1 &2&None \\
			cCIFAR10 &1e-4&256&1e-4&$\{5,1000,5000\}$&0.5& 0.3 &2& Resized crop, horizontal flip \\
			Waterbirds &1e-5&256&1e-4&$\{20,1000,2000\}$&0.1& 0.5 &1.5& Resized crop, horizontal flip \\
			CelebA &1e-5&256&1e-4&$\{20,1000,2000\}$&0.1& 0.5 &1.5& Resized crop, horizontal flip \\
			CivilComments &1e-5&16&1e-4&$\{10,1000,10\text{epoch}\}$&0.1& 0.5 &1.5& None \\
			\bottomrule
		\end{tabular}
	}
\end{table*}

\subsection{Bias-agnostic methods}

To eliminate the dependence on bias annotations/predictions, researchers have developed bias-agnostic methods under different motivations. 
Stable learning based methods \cite{kuang2018stable} propose to decorrelate all features to exclude spurious correlations. For example, \cite{kuang2020stable} 
proposes a sample reweighting approach to decorrelate input variables and demonstrate stable performance across distribution shifts. StableNet \cite{zhang2021deep} further generalizes these techniques to nonlinear feature decorrelation. However, achieving perfect decorrelation theoretically requires a sufficiently large dataset, which is often impractical. To compensate for the imperfectness of sample reweighting under finite-sample settings, Sparse variable independent(SVI) \cite{yu2023stable} introduces a sparsity constraint to avoid unnecessary decorrelation within core variables. Most of these methods are developed within linear frameworks with predefined features, and their extension to deep models is less effective. This is because the learned features from deep models are often heavily entangled, making decorrelation insufficient to disentangle core and spurious factors \cite{yu2023stable}. 
Another line of research utilizes disagreements among diverse models to improve generalization under distribution shifts. For instance, Diversity by disagreement \cite{pagliardiniagree} and Diversify and disambiguate \cite{lee2022diversify} propose to maximize the disagreement between multiple predictors to learn a diverse ensemble, which require a completely spuriously correlated training dataset.
Selective last-layer finetuning (SELF) \cite{labonte2023towards} proposes reweighting the dataset based on misclassifications or disagreements observed before and after last-layer retraining. However, these methods still need a small set of group annotations for model selection. 
Several other bias-agnostic methods have demonstrated competitive performance,
\eg, General Greedy De-bias Learning \cite{han2023general} leverages both a biased model and a base model to assess the difficulty of training samples, thereby focusing training on the more challenging ones. However, this approach may overestimate the level of bias, necessitating the use of curriculum regularization to strike a better balance between in-distribution and out-of-distribution performance.
uLA \cite{tsirigotis2023group} employs biased prediction to adjust the logits for debias learning. It relies heavily on extensively pretrained self-supervised models, which significantly influence the final performance.  
XRM \cite{pezeshki2024discovering} constructs twin networks to fully predict bias across the entire training dataset, using a label-flipping mechanism to intentionally steer the model toward biased predictions.
MASKTUNE \cite{asgari2022masktune}, on the other hand, explores new features during a secondary training phase by masking previously identified spurious factors. However, it requires careful selection of the masking threshold to ensure stable performance. 
The work presented in \cite{10.5555/3618408.3620060} introduces a novel perspective on the causes of spurious correlations. While it shares certain similarities with our approach, it is grounded in a different framework and primarily aims to offer a comprehensive analysis of existing methods.  

Our method also falls within the bias-agnostic framework. 
In contrast to the aforementioned methods, our approach ($i$) does not decorrelate all features, but instead takes a more straightforward approach by maximizing the conditional entropy of label information given the spurious factors;
($ii$) allows for easy control of sample weighting through distribution distance minimization.

\begin{figure}[!h]
	\setlength{\abovecaptionskip}{2.0pt}
	\centering
	\includegraphics[width=0.3\columnwidth]{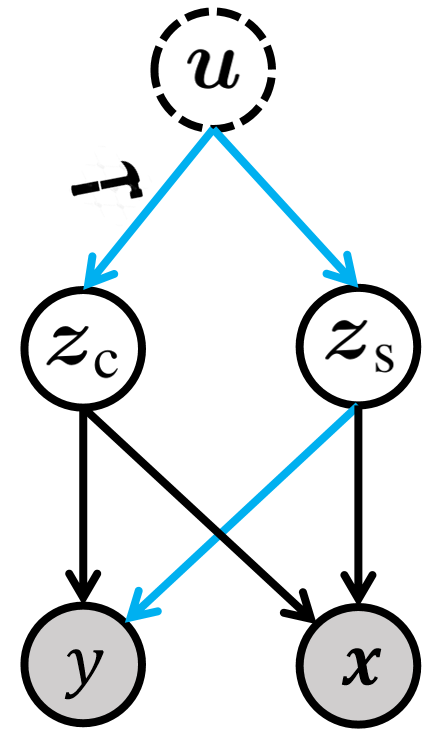}
	\caption{A detailed causal graph. 
	}\label{fig:causal_graph}
\end{figure}

\section{Viewing CCDB from a causal inference perspective}

The causal graph in \cref{fig:three_cases} (left) of the main manuscript is constructed following \cite{neal2020introduction,JMLR:v25:21-1028}. A more detailed version is shown in \cref{fig:causal_graph}, where blue arrows represent paths whose parameters vary across domains, while black arrows denote domain-invariant paths. Here, $\uv$ denotes an unobserved confounder (e.g., the domain or environment) that simultaneously affects both $\zv_{\tc}$ and $\zv_{\ts}$, and this causal influence changes across domains. The mechanism by which $\zv_{\tc}$ and $\zv_{\ts}$ jointly generate the observation $\xv$ is assumed to be invariant across domains. The prediction of $y$ based on $\zv_{\tc}$ is also domain-invariant, whereas the prediction of $y$ based on $\zv_{\ts}$ varies across domains. 
Treating $\zv_{\tc}$ as the treatment and $y$ as the outcome induces a causal path $\zv_{\tc}\rightarrow y$ as well as a backdoor path $\zv_{\tc} \leftarrow \uv \rightarrow \zv_{\ts} \rightarrow y$. 
The spurious correlation between $\zv_{\ts}$ and $y$ arises from the unknown confounder $\uv$.
To safely estimate the causal effect of an image feature $\zv$ on the label $y$, it is necessary to remove the confounding bias induced by the backdoor path and directly estimate the causal effect of $\zv_{\tc}$ on $y$, which remains invariant across domains, namely,

\begin{equation}
	p(y|do(\zv_{\tc})) = \Ebb_{p(\zv_{\ts})}p(y|\zv_{\tc},\zv_{\ts})
\end{equation}
where $do(\zv_{\tc})$ denotes the do-operator from causal inference, which cuts off all incoming paths to $\zv_{\tc}$, thereby removing the spurious correlation between $\zv_{\tc}$ and $\zv_{\ts}$, \ie, $p(\zv_{\ts}|\zv_{\tc}) = p(\zv_{\ts})$, a condition that reflects covariate balance.
Given that the core factor $\zv_{\tc}$ and the label $y$ have a strong one-to-one correspondence, it follows that $p(\zv_{\ts}|y) = p(\zv_{\ts})$. This condition is equivalent to the optimal solution that minimizes the mutual information term in our CCDB framework.

\begin{figure*}[!t]
	\setlength{\abovecaptionskip}{2.0pt}
	\centering
	\includegraphics[width=1.9\columnwidth]{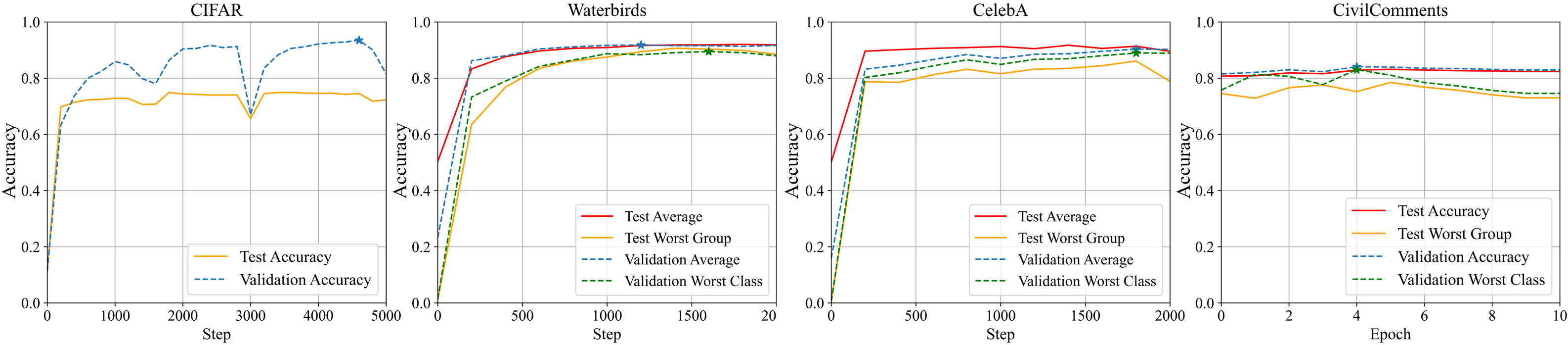}
	\caption{The training process of CCDB on cCIFAR10, Waterbirds, CelebA, and CivilComments. Both the standard average accuracy and the worst-class accuracy on the validation set are effective for selecting models with strong test performance. The best validation accuracy is marked with a star. 
	}\label{fig:train_process}
\end{figure*}

\section{Data}

We conduct experiments on five widely used benchmark datasets: 
cMNIST \cite{li2022invariant}, 
cCIFAR10 \cite{hendrycks2018benchmarking},
Waterbirds \cite{zhou2022model}, 
CelebA \cite{sagawadistributionally}, 
and CivilComments \cite{koh2021wilds} 
to demonstrate the effectiveness of our method.
The detailed statistics of each dataset are summarized in \cref{tab:data_setting}. 
Among these datasets, cMNIST and cCIFAR10 are ten-way classification tasks in which each class is dominated by a specific color or noise pattern, with varying bias ratios.
Waterbirds, CelebA, and CivilComments are real-world datasets: Waterbirds and CelebA are image classification datasets where each class is strongly correlated with a particular background or gender attribute, respectively; 
CivilComments is a text classification dataset where the label is spuriously
correlated with mentions of certain demographic identities.

\section{Training setup}
To make a fair comparison, we apply a 3-hidden layer MLP for cMNIST, a ResNet18 for cCIFAR10, a ResNet50 for Waterbirds and CelebA, and a BERT for CivilComments following existing methods \cite{tsirigotis2023group, liu2021just, pezeshki2024discovering}.
Both ResNet18 and ResNet 50 are pretrained on ImageNet-1K, BERT is pretrained on Book Corpus and English Wikipedia.
The proposed CCDB method consists of three stages, each optimized independently. The first two training stages span 20 epochs and 1000 iterations across all datasets (except CivilComments with 10 epochs); the third stage involves 2000 iterations for Waterbirds and CelebA, 5000 iterations for cMNIST and cCIFAR10, and 10 epochs for CivilComments. 
For Stage1, we use the Adam optimizer with a learning rate of $2\times 10^{-4}$ and default momentum parameters $\beta=(0.9,0.999)$ for all image datasets and AdamW optimizer with a learning rate of $1\times 10^{-5}$ for the NLP dataset; 
For stage2, the Adam optimizer with a learning rate of $1\times 10^{-2}$ is adopted across all datasets; 
The learning rate, along with weight decay and batch size used in stage3 for each dataset are detailed in \cref{tab:augmentation}.

On data preprocessing, we apply simple augmentations to each dataset as listed in \cref{tab:augmentation}, following \cite{ahuja2021invariance}. These augmentations are specifically designed not to induce invariance to variations in the bias attribute. Consequently, the improvements reported in the main paper can be attributed to the proposed CCDB methodology, rather than to the use of bias-nullifying data augmentations.

All the results are averaged over 5 random seeds, with the best-performing model (highest validation accuracy) selected for each seed. 
From the results in \cref{fig:train_process}, we conclude that CCDB can use either standard average accuracy or worst-class accuracy on a class-balanced validation set for model selection.

\begin{figure*}[!t]
	\setlength{\abovecaptionskip}{2.0pt}
	\centering
	\includegraphics[width=1.999\columnwidth]{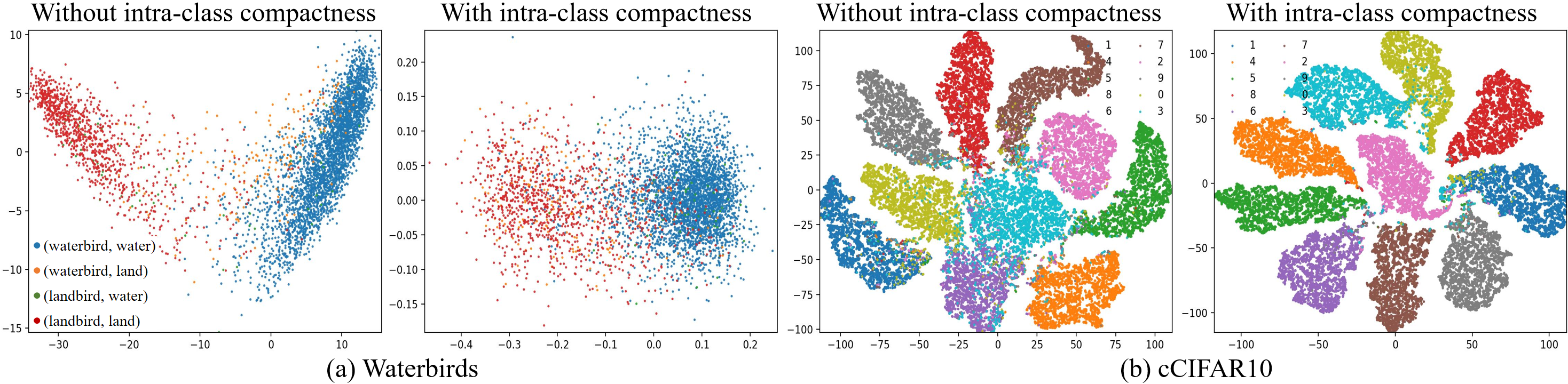}
	\caption{The effect of intra-class compactness on the latent space. t-SNE visualization of the latent space learned on: (a) Waterbirds colored by group annotations; and (b) cCIFAR10 Colored by the spurious noise type attribute. With intra-class compactness, the distribution demonstrates a strong alignment with spurious information.
	}\label{fig:gaussian}
\end{figure*}

\begin{figure*}[!t]
	\setlength{\abovecaptionskip}{2.0pt}
	\centering
	\includegraphics[width=1.999\columnwidth]{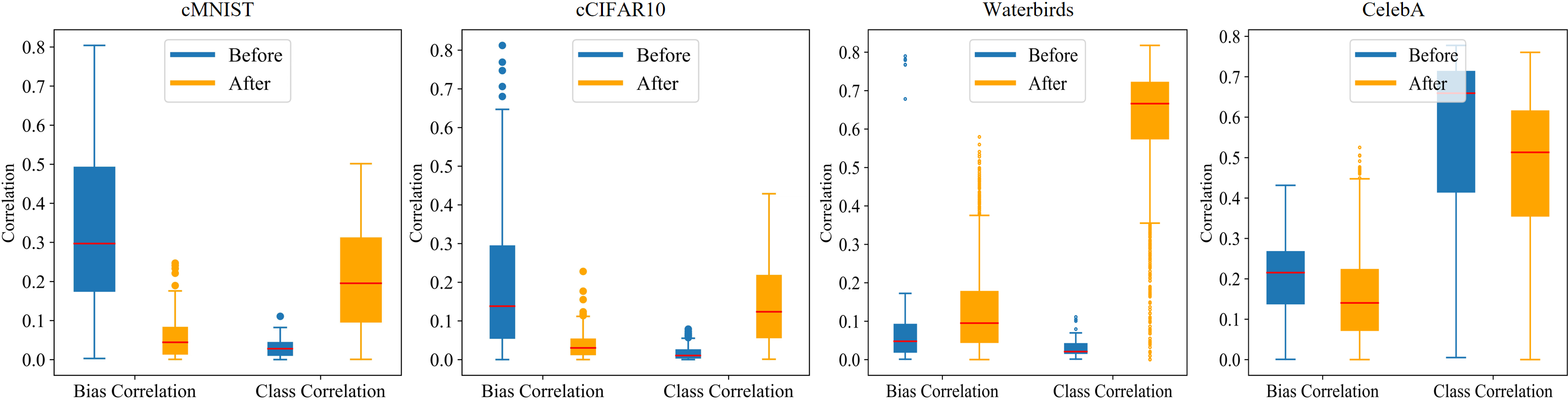}
	\caption{Boxplot of the feature correlations with bias and class information, before and after CCDB sample reweighting.  
	}\label{fig:pearson}
\end{figure*}

\begin{figure}[!t]
	\setlength{\abovecaptionskip}{2.0pt}
	\centering
	\includegraphics[width=0.80\columnwidth]{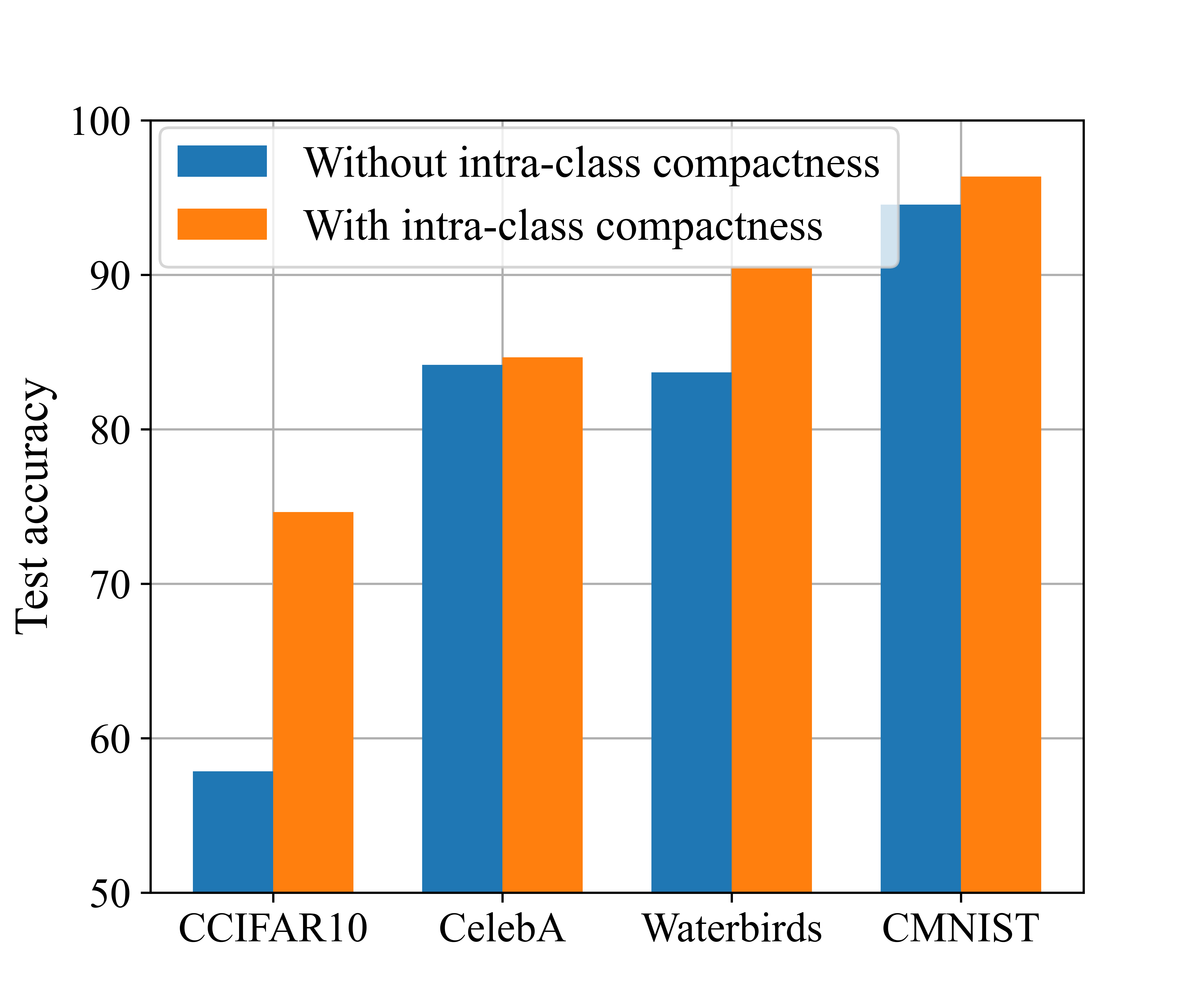}
	\caption{The effect of intra-class compactness on the final performance of CCDB, with the worst group accuracy for Waterbirds and CelebA, and standard average accuracy for cMNIST and cCIFAR10.
	}\label{fig:compact}
	
\end{figure}

\section{Additional experimental results}

\subsection{On model selection for CCDB}

To demonstrate that standard average accuracy on the validation set serves as an effective model selection criterion for CCDB, we record its training process on four datasets (cCIFAR10, Waterbirds, CelebA, and CivilComments), as shown in \cref{fig:train_process}.
The average and worst-group accuracy curves on the validation set exhibit similar trends throughout the training process, with their optimal values occurring in regions where the test worst-group performance remains relatively high. This alignment represents an additional advantage of our method.

\subsection{The computational cost of stage 2}
\label{subsec:stage2}

The stage 2 of CCDB processes all training samples in a single batch, which requires additional storage for the latent representations of the training set. However, the computational cost of this stage is low and acceptable for the following reasons: ($i$) the process operates directly in the latent space, requiring only a tensor of size $N\times D$ to be stored, where $N$ is the number of training samples and $D$ is the dimension of the latent representation. ($ii$) The optimization involves only the calculation of means and covariances, both of which can be performed efficiently.

For optimization of \cref{eq:final_loss}, we use Adam with learning rate $0.01$. \cref{tab:time_stage2} shows the storage and running time on all datasets. Obviously, the optimization is overall fast within a few minutes.

\begin{table}[h]
	\centering
	\caption{
		Computation resource required in Stage2 on four datasets with a single GPU A40. 
	}\label{tab:time_stage2}
	\renewcommand{\arraystretch}{1.4}
	\resizebox{0.99\hsize}{!}{
		\setlength{\tabcolsep}{5pt}
		\begin{tabular}{c|cccc}
			\toprule
			Cost  & cMNIST & cCIFAR10 & Waterbirds & CelebA  \\
			\midrule
			Storage(M) & 6.71 & 22.00 & 2.34  & 79.48   \\
			Time(s) & 80.45 & 114.03 & 39.21 & 232.14   \\
			\bottomrule
		\end{tabular}
	}
\end{table}


\subsection{The effect of intra-class compactness}
The intra-class compactness constraint is applied to amplify bias attributes and smooth the feature space, thereby enhancing the model's generalization ability. In this section, we conduct an ablation study on cMNIST, cCIFAR10, Waterbirds, and CelebA to evaluate the impact of intra-class compactness on the final performance of CCDB.

The results are presented in \cref{fig:gaussian,fig:compact}.
In \cref{fig:gaussian}, under naive ERM training (without the intra-class compactness constraint), the feature distribution is skewed and primarily aligned with class information.
In contrast, when intra-class compactness is applied, the distribution shifts to align more with bias attributes (such as background or noise type). Additionally, the structure of each cluster more closely resembles a Gaussian distribution compared to the results without intra-class compactness.
\cref{fig:compact} shows the average and worst group accuracy with and without the intra-class compactness constraint.
We observe that, for large-scale (\eg, CelebA) and simpler datasets (\eg, cMNIST), the constraint has a relatively modest effect on the final performance. However, for small-scale (\eg, Waterbirds) and complex (\eg, cCIFAR10) datasets, the constraint significantly improves performance, with gains of up to $16\%$ in accuracy.

\subsection{Analysis on sample weights learned by CCDB}


\cref{fig:pearson} shows the effect of sample reweighting on the spuriousness of the learned features. 
We measure spuriousness by calculating the Pearson correlation coefficient \cite{Benesty2009} between each feature dimension and class/bias information, and present a boxplot of these coefficients. 
Note, the coefficients are computed on the test data, where no spurious correlation exists, meaning that class labels are solely correlated with core factors and independent of spurious ones.
Before applying sample reweighting, most extracted feature dimensions from ERM are dominated by spurious factors, exhibiting high correlation with bias information and low correlation with class labels (blue box).
In contrast, after CCDB reweighting, the extracted features are predominantly core, with strong correlation to class labels and weak correlation to bias information (yellow box).
These observations highlight the effectiveness of CCDB in mitigating spurious correlations, thus creating a de-biased data distribution for downstream classification tasks.

\end{document}